\documentclass[10pt,journal,compsoc]{IEEEtran}

\usepackage{times}
\usepackage{epsfig}
\usepackage{graphicx}
\usepackage{amsmath}
\usepackage{amssymb}
\usepackage{color}
\usepackage{subfigure}
\usepackage{comment}
\usepackage{ulem}
\usepackage{booktabs}
\usepackage{multirow,textcomp}

\ifCLASSOPTIONcompsoc
  \usepackage[nocompress]{cite}
\else
  \usepackage{cite}
\fi

\usepackage[breaklinks=true,bookmarks=false]{hyperref}

\newcommand{\norm}[1]{\left\lVert#1\right\rVert}

\begin{document}

\title{Frame Difference-Based Temporal Loss for Video Stylization}

\author{Jianjin~Xu,
        Zheyang~Xiong,
        Xiaolin~Hu,~\IEEEmembership{Senior Member,~IEEE,}
\IEEEcompsocitemizethanks{
\IEEEcompsocthanksitem J. Xu and X. Hu are with the Department of Computer Science and Technology, Tsinghua University, Beijing 100084, China. J. Xu is also with the Department of Computer Science, Columbia University, New York 10027, NY, USA.\protect\\
E-mail: xujj15@gmail.com; xlhu@tsinghua.edu.cn
\IEEEcompsocthanksitem Z. Xiong is with the Department of Computer Science, Rice University, Houston 77005, TX, USA.\protect\\
E-mail: zx21@rice.edu
}
}

\IEEEtitleabstractindextext{
\begin{abstract}
Neural style transfer models have been used to stylize an ordinary video to specific styles. To ensure temporal inconsistency between the frames of the stylized video, a common approach is to estimate the optic flow of the pixels in the original video and make the generated pixels match the estimated optical flow. This is achieved by minimizing an optical flow-based (OFB) loss during model training. However, optical flow estimation is itself a challenging task, particularly in complex scenes. In addition, it incurs a high computational cost. We propose a much simpler temporal loss called the frame difference-based (FDB) loss to solve the temporal inconsistency problem. It is defined as the distance between the difference between the stylized frames and the difference between the original frames. The differences between the two frames are measured in both the pixel space and the feature space specified by the convolutional neural networks. A set of human behavior experiments involving 62 subjects with 25,600 votes showed that the performance of the proposed FDB loss matched that of the OFB loss. The performance was measured by subjective evaluation of stability and stylization quality of the generated videos on two typical video stylization models. The results suggest that the proposed FDB loss is a strong alternative to the commonly used OFB loss for video stylization.
\end{abstract}

}

\maketitle

\section{Introduction}\label{sec:intro}

\IEEEPARstart{I}{n} recent years, neural style transfer for images has achieved great success. Given a style image (e.g.,  ``The Starry Night'' by Vincent van Gogh), these models can transfer an ordinary image to an image with the same style as the style image, which is called {\it stylized image}. The first method reported in this field was an optimization-based algorithm \cite{gatys2016image} that adjusts the input iteratively according to certain goals defined by convolutional neural networks. Subsequently, many computationally more efficient learning-based neural networks were proposed\cite{johnson2016perceptual,ulyanov2016texture,Dumoulin17,LI2017DIVERSIFIED,huang2017arbitrary,chen2017stylebank}.

An interesting extension of these models is to stylize videos. Specifically, given a style image and a video,  a new video is generated for which the frames have a similar style to the style image. Direct generation of video frames using these models will lead to high-frequency flickering artifacts. This deficiency is called {\it temporal incoherence}.

To solve the temporal incoherence problem, most video stylization models track the motion of pixels along time in the original video and use this information to stylize videos. This leads to the need for optical flow \cite{barron1994performance}. Many algorithms have been proposed to estimate optical flow in videos, including (slow) iterative algorithms \cite{barron1994performance,baker2011database} and (fast) deep learning models \cite{dosovitskiy2015flownet,ilg2017flownet}.
Most algorithms are developed under the assumption of {\it brightness constancy}---a pixel's intensity or color does not change when it flows from one frame to another \cite{baker2011database}. This assumption alone leads to an ill-posed problem (one constraint for two unknowns at each pixel), and different algorithms use different prior terms for regularization (see \cite{baker2011database} for a review).

\begin{figure}
     \centering
     \includegraphics[width=1\linewidth]{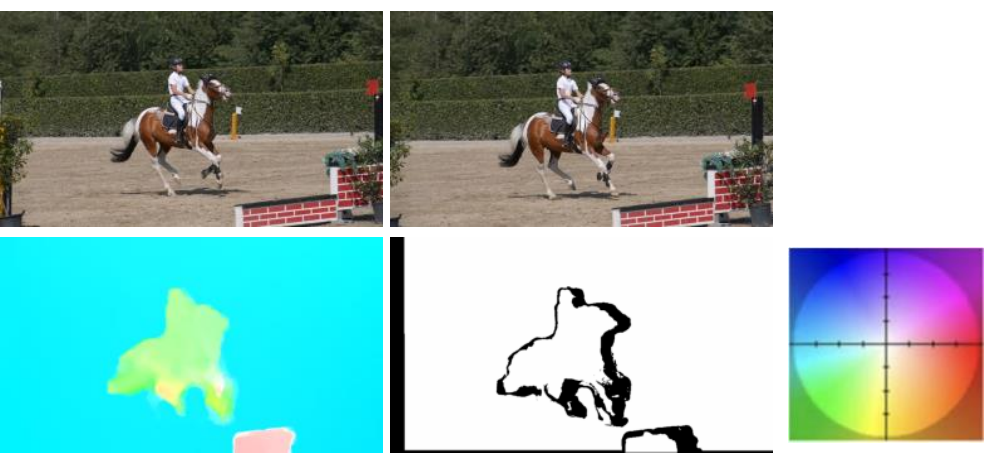}
     \caption{Inaccurate optic flow estimation due to occlusion. Top: two successive frames in a video. Bottom: {\it Left,} the estimated forward optic flow. Different colors indicate different motions, as illustrated on the right. {\it Middle}, the occlusion mask of the optic flow. The value at each pixel is either 0 (black) for the occluded area or 1 (white) for the area that can find a corresponding pixel.}
 \label{fig:ofb_error}
 \end{figure}

It is known that many issues remain unsolved for optical flow estimation in a scene due to effects such as ambient disturbance, illumination change, and occlusion. A perfect algorithm that can solve all of these problems does not exist, and errors are inevitable in certain cases. For example, it is difficult for optical flow algorithms to predict the pixels' movements near the motion boundary due to occlusion. Fig. \ref{fig:ofb_error} shows a person and a horse in the foreground moving to the right (Supplementary Video 1). Since the camera is moving along with the person and the horse, the foreground is roughly static, but the background is moving to the left. The bottom row shows the optical flow results obtained using DeepFlow~\cite{weinzaepfel2013deepflow}. The black regions in the bottom-middle indicate that the estimation near the boundary of the person and the horse has very low confidence. This is because these regions in the background are occluded by the person and the horse in the next frame, and the pixels in these regions do not have corresponding pixels in the next frame. Clearly, this situation worsens if the speed of motion is higher or the frame rate (frames per second) is lower.

Can optical flow be replaced by other methods to stabilize videos during stylization? We reconsider the nature of temporal coherence in videos. Since the original video is assumed to be temporally coherent, the pixel values change in a temporally coherent manner. If the pixel values in the stylized video change in the same manner, then the stylized video should be temporally coherent. We note that the frame difference uniquely determines the temporal evolution of the pixel values in the original video. To achieve temporal consistency,  the difference between two consecutive frames in the generated video can be made  similar to the difference between the corresponding two consecutive frames in the original video. A loss function can be designed in this manner and combined with the original loss functions of any image stylization model to produce temporally coherent videos.

Empirically, we found that this simple method produced excellent stability results but also led to the degradation of stylization quality. We then introduced the frame difference in the feature space specified by a neural network and combined it with the frame difference in the pixel space as the loss function. This loss is called  frame difference-based (FDB) loss. It produced stylized videos with a similar degree of video stability and stylization quality to the optical flow-based (OFB) loss.
Compared with the OFB loss, the proposed FDB loss can significantly reduce the computational cost because it does not estimate optic flow for the training videos before training which typically requires a few days for a normal dataset.

The rest of the paper is organized as follows. In Section \ref{sec:relatedwork}, some closely related works are discussed. In Section \ref{sec:methods}, the FDB loss is formulated and applied to two typical video stylization models. Section \ref{sec:exp} presents the experimental results. Section \ref{sec:conclusions} concludes the paper.

\section{Related Works}\label{sec:relatedwork}
A comprehensive review of image stylization and video stylization is beyond the scope of this paper. We only review some neural network-based methods that are closely related to our method.

\subsection{Neural image stylization}
In 2015, Gatys et al. ~\cite{gatys2016image} proposed a novel method to transfer an image (called {\it content image}) to a new image (called {\it stylized image}) with a style of another image (called {\it style image}). This method is based on the assumption that a pretrained deep neural network for classification on a large dataset can represent the content and style of any image. The basic idea is to fix the neural network and optimize the input such that its content representation and style representation by the neural  network match those of the content image and the style image, respectively. However, optimization is an iterative and time-consuming process. Learning-based methods have been proposed to directly map the content image to the stylized image using a feedforward neural network ~\cite{johnson2016perceptual,ulyanov2016texture}, which are several orders of magnitude faster than the optimization-based method.  A different image stylization method is presented in \cite{kotovenko2019content} to stylize an image according to the style defined by a set of images. One limitation of these methods is that different models  must  be trained for different style images. Several recent methods \cite{Dumoulin17,LI2017DIVERSIFIED,huang2017arbitrary,chen2017stylebank,li2019learning} have been shown to be able to learn multiple styles simultaneously.

\subsection{Neural video stylization}
The critical problem for extending image stylization methods to stylize videos is to ensure temporal coherence in stylized frames.
By introducing a short-term temporal consistency loss based on the optical flow in the frames, it was shown that the optimization-based image stylization model \cite{gatys2016image} can produce stable stylized videos \cite{ruder2016artistic}.  However, these methods inherit the low speed of the optimization-based image stylization method. It is preferred to start from learning-based methods for image stylization \cite{johnson2016perceptual,ulyanov2016texture,Dumoulin17,LI2017DIVERSIFIED,huang2017arbitrary,chen2017stylebank}. A successful attempt of this was the model developed by Huang et al. \cite{huang2017real} that uses the image stylization model of Johnson et al.\cite{johnson2016perceptual} but introduces a short-term temporal loss for training the generative network. Another successful attempt was carried out by Ruder et al.'s \cite{ruder2018artistic} who developed a model that also uses the  image stylization model of Johnson et al.\cite{johnson2016perceptual} but introduces both short-term temporal loss and long-term temporal loss. A recurrent neural network was proposed to stabilize the generated video frames~\cite{gupta2017characterizing}. This approach differs from that of the model of Huang et al.~\cite{huang2017real} in the generation process, in which the output of the model at the previous time step is needed as input to the model to generate the output at the current time step. With this setting, the original video frames are required to be input to the model sequentially during training and testing. A patch-based image stylization method \cite{frigo2016split} is extended to stylize videos \cite{frigo2019video}. Image parts are tracked in different frames based on an optical flow method and inconsistent patches are synthesized. Another method estimates a low-resolution optic flow in every training step and minimizes a compound temporal loss to stabilize the videos \cite{wang2020relax}.

To ensure temporal coherence among video frames, the above methods usually employ OFB temporal losses to train the stylization networks. After training, the stylization networks are different from those trained without any temporal loss. The stylized frames in the two settings are different; in other words, the temporal loss may degrade the stylization quality of the individual frames. Chen et al. \cite{chen2017coherent} proposed to fix the stylization network but learn to make use of the previous frame to avoid inconsistency. Each new frame is obtained by linearly combining two frames with a learnable mask. Both frames are the outputs of the stylization network, one from the current time step and the other from the last time step but are warped to the current time step using a feature level warp function based on optical flow. This method is quite different from the existing methods and from the method proposed in this paper because it does not change the stylization network.

\subsection{Temporal consistency}
In addition to video stylization, temporal consistency must be achieved in any video editing application. Optical flow is also widely used in these applications. Kong et al. \cite{kong2014intrinsic} proposed to extract temporally coherent albedo and shading from the video (intrinsic video) using optical flow. In addition, it was shown that intrinsic video was useful to improve optical flow estimation. Two different methods \cite{Ye14,bonneel2014interactive} were proposed for intrinsic video decomposition, where the temporal consistency was ensured by using optical flow. Litwinowicz et al. \cite{litwinowicz1997processing} proposed to transform ordinary video into animations that had an impressionist effect. Optical flow fields were used to push brush strokes from frame to frame in a temporally consistent manner. Farbman and Lischinski \cite{farbman2011tonal} proposed to stabilize tonal fluctuations in videos by using optical flow estimated at sparse locations. Bonneel et al. \cite{bonneel2015blind} proposed to stabilize many image processing tasks, including stylization of color and tone and depth estimation using dense optical flow. Lai et al. ~\cite{lai2018learning} proposed a cascaded solution to address the video stability problem in a wide range of tasks by training an RNN using OFB temporal consistency loss. Most of these methods consider the local temporal consistency between the adjacent frames explicitly.
A generative adversarial network-based method was proposed to enhance the global consistency in many frames \cite{wei2018video}.

Since optical flow must be estimated in every frame, its estimation also faces the stability problem. Many methods have been proposed for stabilizing the optical flow across frames, including spatiotemporal filtering \cite{hosni2011temporally} and temporal edge-aware filtering \cite{lang2012practical}.

 In addition to optical flow, several other techniques can also stabilize frames in video editing. Motivated by Gaussian convolution, Paris \cite{paris2008edge} devised an edge-preserving smoothing technique. Meka et al. \cite{meka2016live} proposed a global, sampling-based, and spatiotemporal reflectance consistency constraint. Bonneel et al. \cite{bonneel2013example} proposed a method based on high-dimensional differential geometry. However, most of these methods are problem-specific and cannot be directly used for neural video stylization.

\section{Methods}\label{sec:methods}
We consider a class of video stylization models \cite{huang2017real,gupta2017characterizing} that uses a loss for individual frames (called {\it spatial loss} in what follows) and a loss for between-frames (called {\it temporal loss}).  Our contribution is for a new temporal loss, but to make the method self-contained, we briefly review the spatial loss first.

\subsection{Spatial loss}
The spatial loss usually has two parts, content loss and style loss~\cite{gatys2016image}. The former is focused on preserving the content of the content image, and the latter is focused on transferring the style to the style image. Current learning-based neural networks for image style transfer usually rely on a neural network called the {\it loss network} the parameters of which are pretrained on other tasks (e.g., classification on the ImageNet dataset) and fixed. Both content loss and style loss are computed on the top of this loss network and are used to train a stylization network  d parameters need to be learned.

We denote the content image by $I$, the style image by $S$ and the stylized image (output) by $\tilde I$. Let $f_l(x)$ be the feature map of layer $l$ with $x$ being the input to the pretrained loss network, specifically, either $I$ or $\tilde I$. For simplicity, batch size is set to 1. Then, a typical neural network feature map has a 3D shape $C_l\times H_l\times W_l$, where $C_l$ denotes the number of channels, and $H_l$ and $W_l$ denote the height and width. The feature map $f_l(x)$ can be flattened into a 2D matrix $F_l(x)$ with the shape $C_l \times H_l W_l$. The Gram matrix of $f_l(x)$ is defined as~\cite{gatys2016image}:
\begin{equation*}
    G_l(x) = f_l(x) f_l(x)^\top.
\end{equation*}

The content loss in a layer $l$ is defined as the Euclidean distance between $I$ and $\tilde I$ at the feature level
\begin{equation}\label{eq:cont-layer}
    E_\text{cont}^{l}(\tilde I, I) = \frac{1}{2 C_l H_l W_l} \norm{ f_l(\tilde I) - f_l(I) }^2_2.
\end{equation}
The style loss in a layer $l$ is defined as the Euclidean distance between the Gram matrices of the feature maps of $S$ and $\tilde I$
\begin{equation}\label{eq:sty-layer1}
    E_\text{sty}^{l}(\tilde I, S) = \frac{1}{2 C_l^2} \norm{ G_l(\tilde I) - G_l(S) }^2_2.
\end{equation}
This loss is proposed in \cite{gatys2016image} and has been widely used in neural image stylization \cite{johnson2016perceptual,ulyanov2016texture,Dumoulin17}.

The content losses and  style losses from different layers are weighted and summed together to yield the final losses
\begin{equation}\label{eq:contloss}
    L_\text{cont}(\tilde I, I) = \sum_{l \in J_c} a_l E_\text{cont}^l(\tilde I, I)
\end{equation}
\begin{equation}\label{eq:styloss}
    L_\text{sty}(\tilde I, S) = \sum_{l \in J_s} b_l E_\text{sty}^l(\tilde I, S)
\end{equation}
where $J_c$ and $J_s$ denote the sets of layers chosen to calculate the content losses and style losses, respectively, and $a_l$ and $b_l$ denote the weighting factors.

\subsection{Temporal loss} \label{sec:temploss}

Since the original video (or {\it content video}) is assumed to be temporally coherent, and this coherence is encoded in the difference between frames, by requiring the stylized video to have a similar between-frame difference in a model, one may obtain stabilized video output.

We denote the original frames by $\{I_t\}$ and the stylized frames by $\{\tilde I_t\}$. The temporal loss is defined as
\begin{equation}\label{eq:temploss}
    L_\text{temp}(\{\tilde I_t\}, \{I_t\}) = \frac{1}{2N(T-1)} \sum_{t=1}^{T-1}  \norm{\phi(\tilde I_t) - \phi(I_t)}_2^2,
\end{equation}
where $N$ is the number of pixels in each frame, $T$ is the total number of frames, and
\begin{equation}\label{eq:framediff}
\phi(x_t) = f_l(x_{t+1}) - f_l(x_{t}),
\end{equation}
where $t=1,\ldots,T-1$, is the frame difference between two consecutive frames $x_{t+1}$ and $x_{t}$. If $l=0$, $\phi(x_t)$ is simply the frame difference in the pixel space
\begin{equation}\label{eq:framediff-pix}
\phi(x_t) = x_{t+1} - x_{t}.
\end{equation}
The temporal loss \eqref{eq:temploss} with the frame difference defined in the pixel space ($l=0$) is called the {\it pixel-FDB (P-FDB) loss}, denoted by $L_\text{pt}$. If $l>0$ in \eqref{eq:framediff}, then the frame difference is measured in the feature space, and the loss is called the {\it feature-FDB (F-FDB) loss} and denoted by $L_\text{ft}$. The P-FDB and F-FDB can be weighted and summed to form the {\it combined-FDB (C-FDB) loss}, denoted by $L_\text{ct}$.

We would like to remark that measuring the difference between consecutive frames is not the only choice for $\phi(x_t)$. In fact, the difference between any pair of frames $x_{t+K}$ and $x_{t}$ ($1\le K\le T-1$) can be used, and their performances are similar (see Section \ref{sec:pixel-FDB} for empirical results).

For any neural video stylization model, we can define the total loss for minimization:
\begin{equation}\label{eq:totalloss}
L = \lambda_1 L_\text{cont} + \lambda_2 L_\text{sty} + \lambda_3 L_\text{temp},
\end{equation}
where $\lambda_i$'s are the weighting factors. We note that $L_\text{temp}$ can be $L_\text{pt}$, $L_\text{ft}$ or $L_\text{ct}$.

\subsection{Application to video stylization models}\label{sec:SFN}

\begin{figure*}
    \centering
    \subfigure[The SFN with the P-FDB loss.]{\label{fig:SFNa}
        \includegraphics[width=0.48\linewidth]{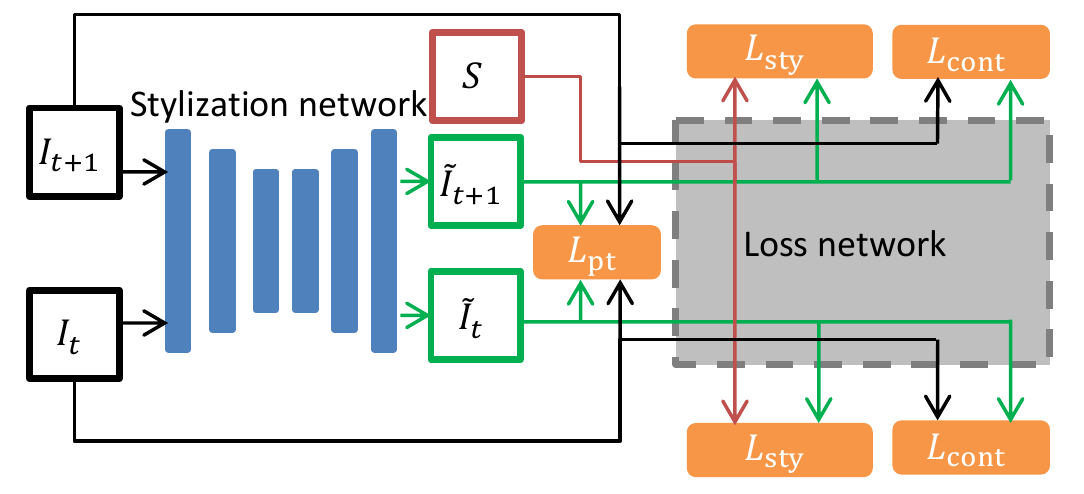}
    }
    \subfigure[The SFN with the F-FDB loss.]{\label{fig:SFNb}
        \includegraphics[width=0.48\linewidth]{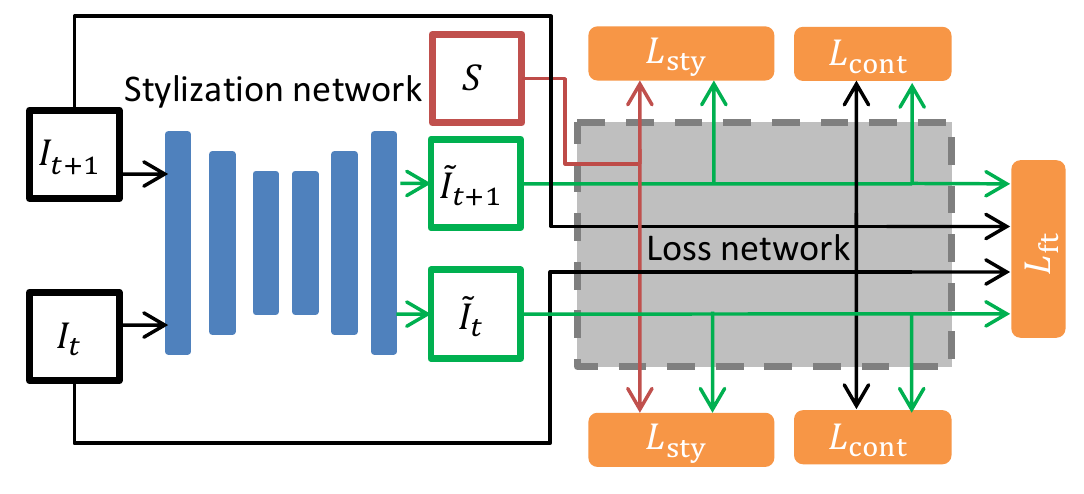}
    }
    \subfigure[The RNN with the P-FDB loss.]{\label{fig:RNNa}
        \includegraphics[width=0.48\linewidth]{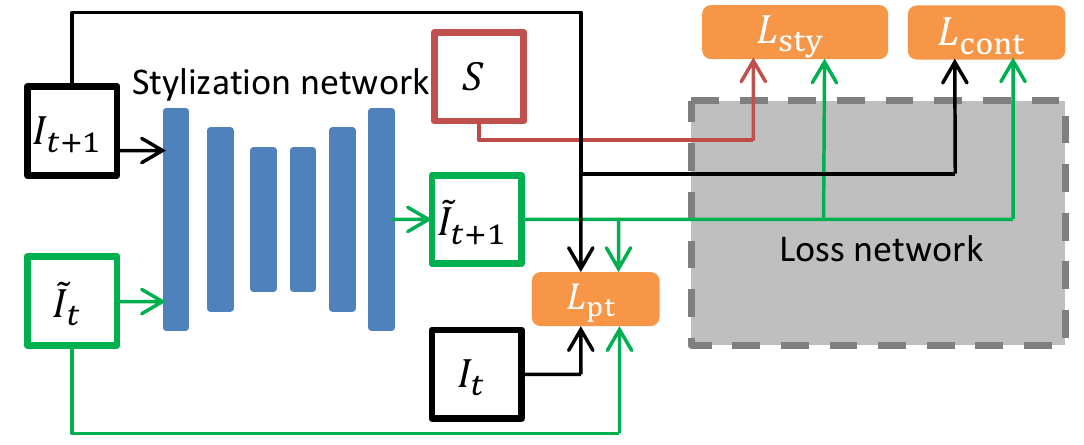}
    }
    \subfigure[The RNN with the F-FDB loss.]{\label{fig:RNNb}
        \includegraphics[width=0.48\linewidth]{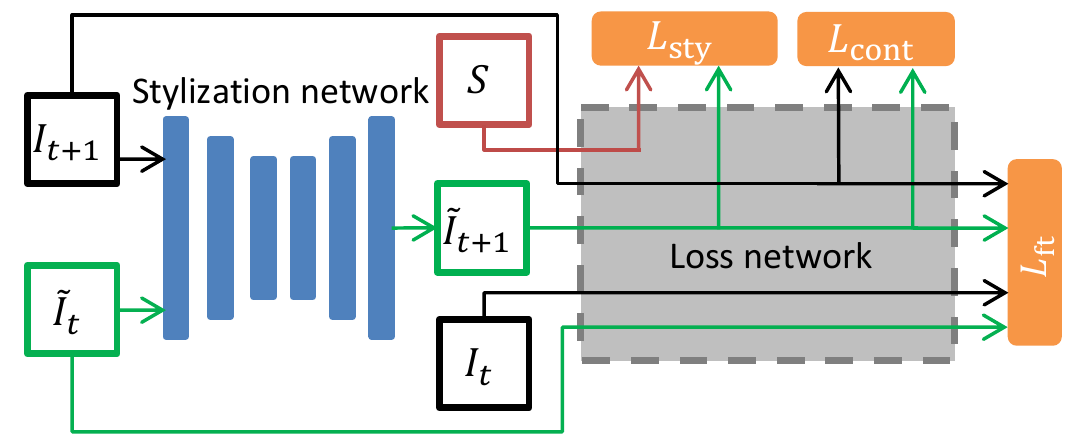}
    }
    \caption{Training pipelines for the SFN and the RNN. The P-FDB loss and the F-FDB loss are defined in \eqref{eq:framediff} and  correspond to $l = 0$ and $l > 0$, respectively. We note that in either case, the temporal loss needs four inputs: $I_t, I_{t+1}, \tilde{I}_t$ and $\tilde{I}_{t+1}$. For the SFN, in inference, at every time step $t$, only one frame $I_t$ from the original video is used and a stylized frame $\tilde I_t$ is output. For the RNN, in both training and inference, previous stylization output $\tilde{I}_t$ (or zero at the first frame) and the current frame $I_{t+1}$ are used to produce stylized output $\tilde{I}_{t+1}$.}
    \label{fig:pipeline}
\end{figure*}

Generally, the proposed FDB loss can be applied to any video stylization model that uses the OFB loss. In this paper, we present its application to two typical models:
\begin{itemize}
\item A simple yet efficient model proposed in~\cite{huang2017real}. We refer to this  as Simple Feedforward Network (SFN).
\item A Recurrent Neural Network (RNN)  proposed in~\cite{gupta2017characterizing}.
\end{itemize}
Both models consist of a stylization network and a loss network. The stylization networks are used to generate video frames and the loss networks that are usually pretrained on the ImageNet classification task and then fixed are used to calculate the losses for training the stylization networks.
For both models, we substitute the original OFB loss with the FDB loss and keep other losses the same.

During training, the SFN takes in two successive frames to learn the temporal coherence. The training pipelines with the P-FDB loss and the F-FDB loss are slightly different, as illustrated in Figs.~\ref{fig:SFNa} and ~\ref{fig:SFNb}, respectively. We note that the two frames form a minibatch with size 2 and are not concatenated along the channel axis. Let the successive frames be $\{I_{t}, I_{t+1}\}$ and their stylized outputs be $\{\tilde I_{t}, \tilde I_{t+1}\}$. Then, the temporal loss in \eqref{eq:temploss} is properly defined, and the total loss \eqref{eq:totalloss} is used to optimize the stylization network. We note that there are two style losses for each pair of inputs, two content losses, and just one temporal loss.

For comparison, we can replace the FDB temporal loss with the OFB temporal loss to transform the SFN back to the model in \cite{huang2017real}. A typical form of the OFB loss is
\begin{equation}\label{eq:ofbloss}
    L_\text{temp} = \frac{1}{2N(T-1)} \sum_{t=1}^{T-1}  \norm{c_k(\tilde I_{t+1}-\omega(\tilde I_{t}))}_2^2,
\end{equation}
where $\omega$ is a function that warps the stylized output at time $t$ to time $t+1$ according to the optical flow, and $c\in [0,1]^{N}$ denotes the per-pixel confidence of the optical flow.
Readers can refer to \cite{ruder2016artistic,huang2017real,gupta2017characterizing} for the details of the OFB loss.

We note that during inference, the SFN takes one frame as input and outputs a stylized frame, and the loss network is no longer needed.

The training pipelines for the RNN with the P-FDB loss and the F-FDB loss are shown in Figs.~\ref{fig:RNNa} and \ref{fig:RNNb}, respectively. The RNN takes in two images together, namely, the previous stylized frame $\tilde I_t$ and the current content frame $I_{t+1}$. These  are concatenated along the channel axis and then fed into the RNN. Except for the difference in the input layer, the other layers of the RNN are the same as the SFN. Since the ``previous'' stylization frame of the first frame does not exist, it is set to a matrix of all zeros. The temporal loss \eqref{eq:temploss} is calculated based on $I_{t}, I_{t+1},\tilde I_{t}$ and $\tilde I_{t+1}$.
In training and inference, the model works in the same manner. Replacing the FDB loss with the OFB loss transforms the model back to the model in \cite{gupta2017characterizing}.

\section{Experiments}\label{sec:exp}

\subsection{Dataset}\label{sec:dataset}
The training split of DAVIS~\cite{Perazzi2016} was used as the training set. DAVIS is a real-world video dataset covering several difficult scenarios such as fast motion, motion blur, and occlusion. It consists of 50 video clips and approximately 3,000 frames. No preprocessing was needed for the FDB loss, while for the OFB loss, we estimated the optical flow for the entire dataset prior to training. DeepFlow~\cite{weinzaepfel2013deepflow} was adopted for estimating the optic flow, requiring the computational time of approximately 3 days.

Our test set consisted of randomly selected ten videos from the DAVIS test split and ten videos from the Sintel~\cite{Butler:ECCV:2012} test split. Sintel is a computer-generated movie dataset and is widely used in neural stylization studies.

\subsection{Implementation details}\label{subsec:implementation}
We used the same architecture as~\cite{huang2017real} for the SFN but used a new padding approach in every convolutional layer. We note that padding is necessary to keep the size of the stylized image the same as that of the original image due to the convolution operations in the model that shrink the size of the input image. This network does not have pooling layers but has convolutional layers with stride larger than 1. Even though upsampling layers are present, without appropriate padding, the network still outputs smaller images than the input images. In experiments, we found that the ordinary padding approaches were prone to produce bounding box artifacts (horizontal or vertical bars appearing around the boundary of the stylized video), even if no temporal loss was applied. We found that this artifact could be largely suppressed by adopting an interpolation padding approach. First, the feature maps with spatial size $H\times W$ are bilinearly upsampled to the desired size $H'\times W'$. Then, the interior parts with the spatial sizes of $H\times W$ are substituted by original feature maps. In other words, the padded area is obtained by bilinear interpolation. This approach was adopted in both the SFN and the RNN in all experiments.  Its effect is demonstrated in Section ~\ref{sec:padding}.

The loss network was a modified version of VGG-16 \cite{simonyan2014very} (all max-pooling layers were replaced with average pooling layers as suggested in \cite{gatys2016image}), with its 16th layer as the content layer and the 4th, 9th, 16th, 23rd layers as the style layers. For comparison, we trained three SFN models with the P-FDB loss, the OFB loss, and without any temporal loss, respectively. We tried our best to make each model achieve its best performance by tuning the hyper-parameters.

 The SFN with any temporal loss was trained in two steps. First, it was trained as an image stylization model as described in \cite{johnson2016perceptual}. We used the Adam~\cite{kingma2014adam} optimizer with a learning rate of $10^{-3}$ and batch size 4 and optimized the model for 20k iterations. Second, the SFN was finetuned on the DAVIS training set with a certain temporal loss. The weights of the content loss and the style loss were kept the same as in the first step to ensure that the stylization did not change. The finetuning process required two epochs (3456 iterations) using the Adam optimizer with a learning rate of $10^{-4}$ and batch size 2 (a pair of successive frames). We observed that two epochs of training were sufficient to stabilize the videos. The first step took approximately 6 hours. The training time for the second step depended on the temporal loss but was always approximately 20 minutes. Without the temporal loss, we only trained the SFN in the first step.

The RNN with any temporal loss  was trained followed the setup in~\cite{gupta2017characterizing}. First, the RNN was initialized with the SFN trained in the first step. The first convolutional layer of the RNN that had six input channels was learned from scratch. Second, the RNN was finetuned with a batch size of 4 according to~\cite{gupta2017characterizing}. Four consecutive frames formed a minibatch. The other training settings were the same as the SFN. The finetuning step took approximately one hour for different temporal losses.

For all models, we set $\lambda_1=1$,  $\lambda_2=10$ and $\lambda_3=400$. For the C-FDB loss, we set $L_\text{ct}=\frac{1}{40} L_\text{pt} + L_\text{ft}$. Therefore, $\lambda_3 L_\text{ct} = 10 L_\text{pt} + 400 L_\text{ft}$. The 9-th layer of VGG16 ($l=9$ in \eqref{eq:framediff}) was used  to calculate the frame difference in the feature space.

All of the experiments were conducted on a single NVIDIA GeForce GTX TITAN X  GPU. All models were implemented in Pytorch~\cite{NEURIPS2019_9015} and the source codes are publicly available\footnote{\url{https://github.com/AtlantixJJ/frame-difference-loss}}.

\subsection{Evaluation method}\label{sec:eval}
We evaluate the video stability and frame stylization quality for models trained with different temporal losses. It is difficult to find an objective metric to measure perceptual stability and quality.
Some widely used video quality metrics, including PSNR and SSIM, do not accurately reflect human perception. These metrics fail to account for many nuances of human perception~\cite{Zhang2018unreasonable} that are important for the perception of the flicker artifacts in videos. Therefore, we carried out behavioral experiments to evaluate the proposed FDB loss. The study was approved by the Department of Psychology Ethics Committee, Tsinghua University, Beijing, China.

We adopted the classical two-alternative forced choice~(2AFC) experimental paradigm. To evaluate video stability, pairs of videos were presented to the subject simultaneously, and the subject was instructed to pick the more stable video. Considering that a subject may not be familiar with what stability means, we collected a few examples from demo videos (a stable video and an unstable video) from previous video stylization studies \cite{ruder2016artistic,chen2017coherent}. To evaluate frame stylization quality, pairs of frames along with the corresponding style image were presented to the subject, and the subject was instructed to select the image based on the similarity of its style to the reference style image. The contents of videos or frames and the style images were controlled to be the same for comparing the performances of two different temporal losses.

\begin{figure}
	\centering
    \includegraphics[width=0.8\linewidth]{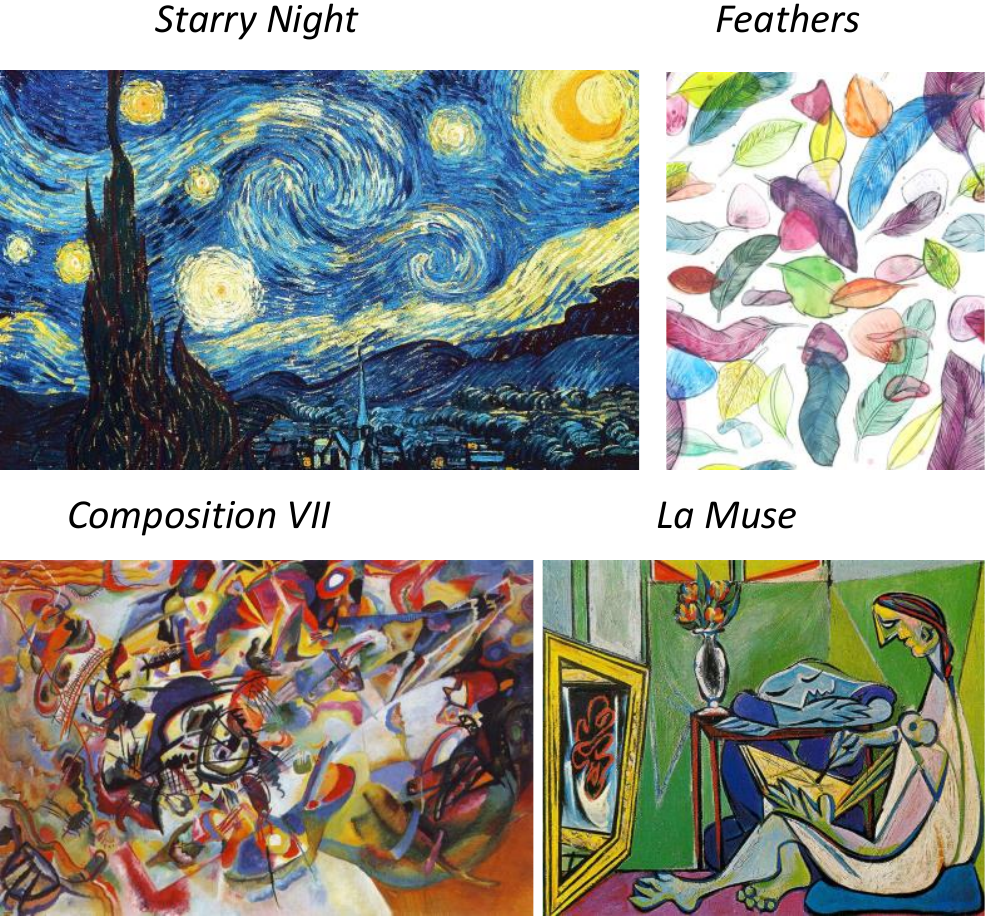}
    \caption{Four style images used in this study.}
    \label{fig:styleimage}
\end{figure}

To evaluate video stability, we stylized each test video into four videos with style images, ``Starry Night'', ``La Muse'', ``Feathers'' and ``Composition VII'' (Fig. \ref{fig:styleimage}).
Since there were 20 videos in the test set, every method generated 80 videos for comparison. To evaluate frame stylization quality, we randomly chose frames from the 80 stylized videos and ensured that the frame indices were the same for different temporal losses to be compared. The order of trials was randomly shuffled but fixed for all subjects.

For the comparison of a pair of losses (e.g., the P-FDB loss and the OFB loss), one subject must finish two sets of evaluations: one set for video stability and the other for frame stylization quality. This is called a \textit{session}. In other words, in every session, a subject should make 80 votes for video stability and 80 votes for frame stylization quality. A subject must complete all of the 160 votes before she or he can participate in another session for comparing the results with another pair of losses. Each session was assigned to 20 different subjects, resulting in $20 \times 160 = 3,200$ votes. There were eight different sessions (three in Table \ref{Tab:pixelFDB}, three in Table \ref{Tab:combinedFDB} and two in Table \ref{Tab:RNN}). In total, we collected $8 \times 20 \times 160 = 25,600$ votes.

We recruited 62 subjects (31 male and 31 female, aged between 18 to 30). They were paid according to the number of sessions completed. Among the subjects, 27 of them had normal vision. The others had corrected-to-normal vision.

We performed most analyses based on the SFN and present the results in Sections~\ref{sec:padding} to \ref{sec:combinedFDB}). We present the main results of the RNN in Section~\ref{sec:exp-RNN}.

\subsection{Comparison of padding approaches}\label{sec:padding}

\begin{figure*}
    \centering
    \includegraphics[width=1.0\linewidth]{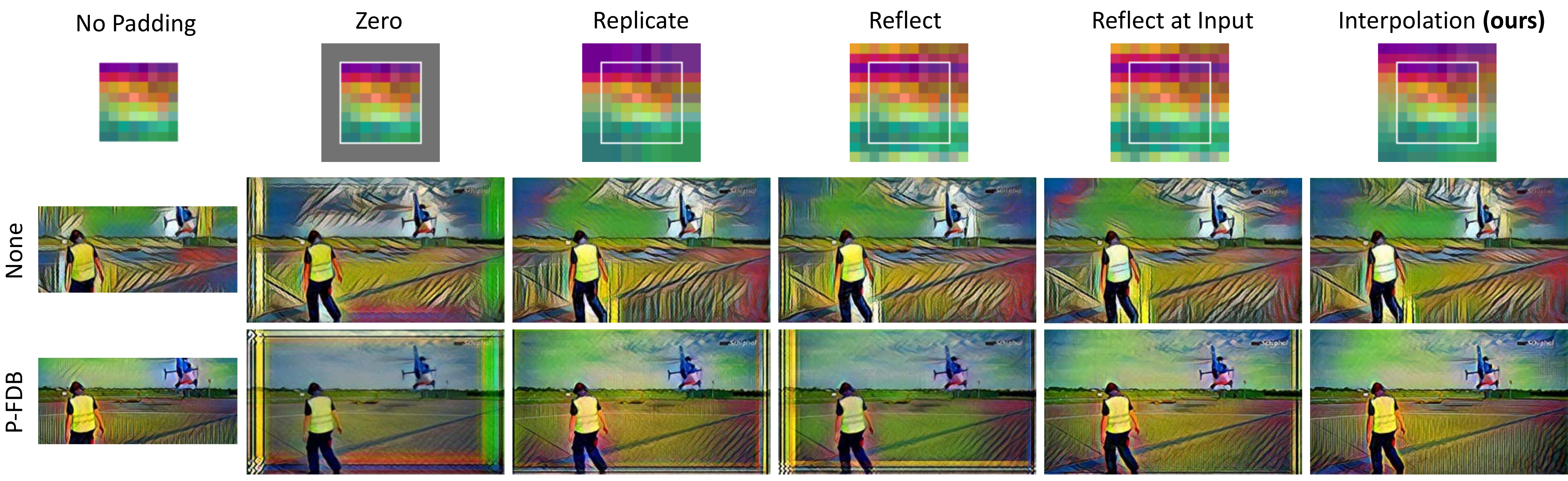}
    \caption{Results of the SFN models using different padding approaches. The first row illustrates various padding approaches. The second and the third rows show the results of the SFN trained without temporal loss and with the P-FDB loss, respectively. The style image is ``La Muse''.}
    \label{fig:padding_comparison}
\end{figure*}

First, we found that if convolutions were used without padding, the bounding box artifacts mentioned in Section \ref{subsec:implementation} did not appear, but the stylized videos had smaller spatial sizes than the original videos  (Fig. ~\ref{fig:padding_comparison}, leftmost column). As discussed in Section \ref{subsec:implementation}, to keep the spatial size of the original video, padding is necessary. We tried different padding approaches (Fig. ~\ref{fig:padding_comparison}), namely:
\begin{itemize}
\item Zero padding: put zeros for the surrounding of the feature maps;
\item Replicate padding: put the values of the border for the surrounding of the feature maps;
\item Reflective padding: reflect every feature map to its surrounding as if there was a mirror on each of the four borders;
\item Reflective-at-input padding\footnote{Used in implementing the image stylization network in \cite{johnson2016perceptual}. See \url{https://github.com/jcjohnson/fast-neural-style}}: same as reflective padding, but it is only used in the input image, while the convolutional layers do not use any padding.
\end{itemize}
All of these approaches were prone to produce bounding box artifacts. Specifically, for the zero padding approach and the reflective padding approach, the artifacts appeared even without temporal loss. For the replicate padding approach and the reflective-at-input padding approach, the artifacts did not appear in the results of the SFN trained without the temporal loss but appeared in the results of the SFN trained with the P-FDB loss. Instead, our proposed interpolation padding approach as described in Section \ref{subsec:implementation} did not  produce the bounding box artifacts.

\subsection{Results with the P-FDB loss}\label{sec:pixel-FDB}

We first studied the FDB loss defined in the pixel space. The training pipeline of the SFN is illustrated in Fig.~\ref{fig:SFNa}.

\subsubsection{Video stability: qualitative results} \label{sec:pixel-stability-qualitative}

First, we compared the results of the SFNs trained with and without temporal losses qualitatively.  Without any temporal loss, the videos produced by the SFN had flickering effects (see Supplementary Video 2 and Fig.~\ref{fig:sfn_diff_flow_none}). In row ``None'' in Fig.~\ref{fig:sfn_diff_flow_none}, the zoom-in red box (upper part) demonstrates that the dark blue textures on the bamboos in frame 1 change to sharp yellow stripes in frame 2. In the blue box of the other pair of frames, the number and mode of stripes of the two frames do not match\footnote{In our experiments, we found difficult to demonstrate the instability of a video by comparing consecutive frames, because the flickering effect usually had a high spatial frequency. Readers are invited to watch the companion supplementary videos as indicated in the captions of figures.}. By contrast, both the P-FDB loss and the OFB loss can stabilize the stylized videos. The bottom two rows in Fig.~\ref{fig:sfn_diff_flow_none} show higher consistency between two successive frames generated by the SFNs with the OFB loss and the P-FDB loss.

\begin{figure*}
    \centering
    \includegraphics[width=0.99\linewidth]{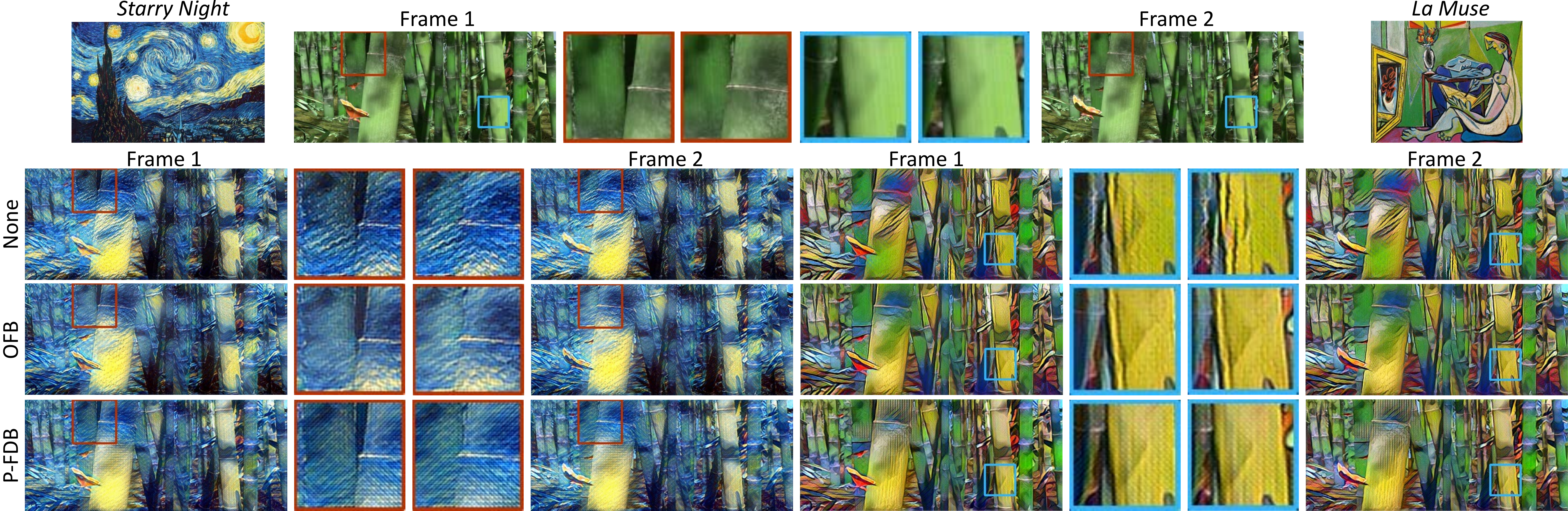}
    \caption{Results of three SFNs given a content video (``bamboo\_3" in the Sintel dataset) and two style images (``Starry Night'' and ``La Muse''). The first row shows the two style images and two successive frames in the content video. Zoomed-in images are also shown as indicated by the red and blue boxes. The second to the fourth rows show the stylized frames obtained by the SFNs trained without temporal loss, with the OFB loss, and the P-FDB loss, respectively. See Supplementary Video 2. }
    \label{fig:sfn_diff_flow_none}
\end{figure*}

\begin{figure}
    \centering
    \includegraphics[width=\linewidth]{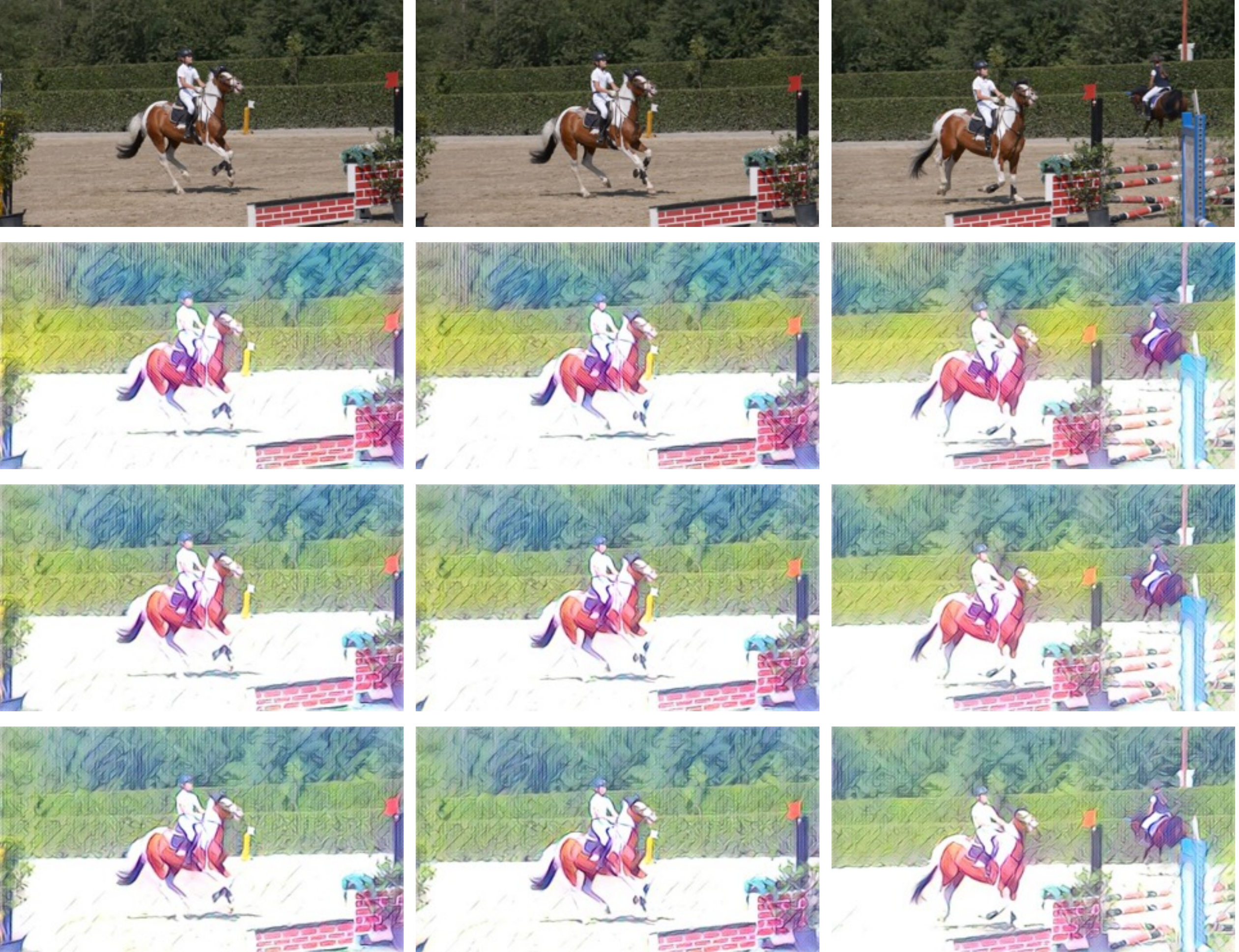}
    \caption{Sample results of the SFNs with different frame intervals. The models were trained with the style image ``Feathers''. The first row shows three original frames (No. 0, 1, and 9) from a clip in the DAVIS test split. The second to the fourth rows show the stylized frames by SFN trained with the 1, 4, and 8 frame interval P-FDB losses, respectively. See Supplementary Video 3. }
    \label{fig:frameinterval}
\end{figure}

Second, we tested the approach of using a larger frame interval in the frame difference measurement. As explained in Section~\ref{sec:temploss}, using the difference between two frames that are separated by an interval of $K$ frames  where $K>1$ should work as well as using $K=1$. This was verified by the experiments. As shown in Fig.~\ref{fig:frameinterval}, the stylization results of models trained with different $K$ are almost identical to each other. See Supplementary Video 3.

Third, we found that the models trained with the P-FDB loss were able to maintain long-term consistency to some extent. An example is shown in Supplementary Video 4 and Fig.~\ref{fig:occlusion_comparation}. When the camera moved from left to right beside a pillar in a scene, a part of the scene was occluded by the pillar for some time and then reappeared. After reappearing, the regions before occlusion were stylized in a consistent manner. By contrast, in the absence of the P-FDB loss, the stripes in the red box before occlusion disappeared after this region reappeared. We note that unlike a previous study \cite{ruder2016artistic}, we did not impose any explicit mechanism on the proposed model to ensure long-term consistency. We speculate that this long-term consistency is a result of learned resistance to the disturbance of input: if the input stays the same, the stylized output is also trained to stay the same.

\begin{figure}
	\centering
    \includegraphics[width=1\linewidth]{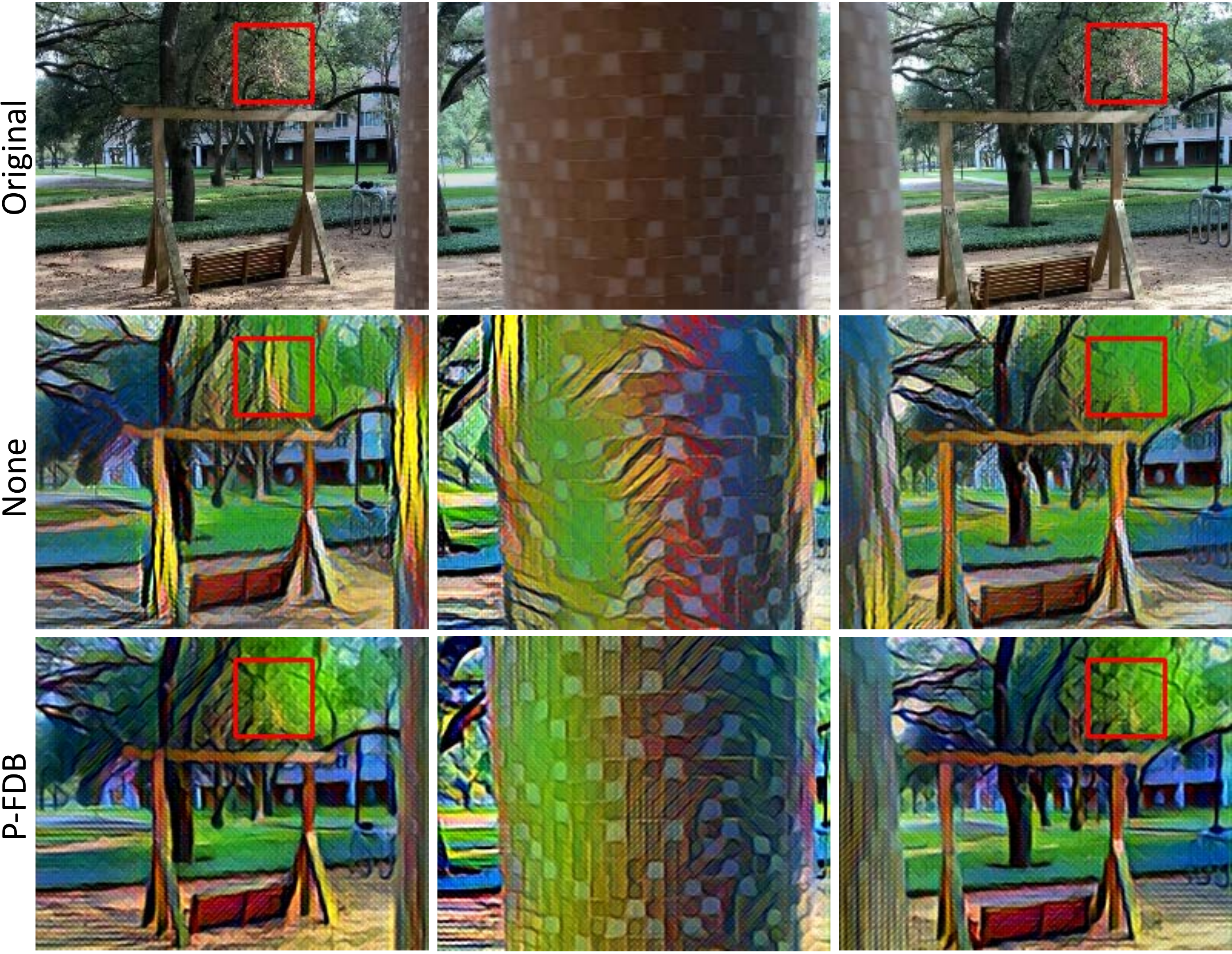}
    \caption{An example of long-term consistency in the models trained with the P-FDB loss. The first row shows three frames (frame interval is 33 and 23 respectively) in an original video recorded in a park. The second and third rows show the stylized frames obtained by SFNs trained without temporal loss and with the P-FDB loss, respectively. The style image is ``La Muse''. See Supplementary Video 4.  }
    \label{fig:occlusion_comparation}
\end{figure}

\subsubsection{Video stability: quantitative results} \label{sec:pixel-stability-quantitative}

Throughout the paper, if two methods A and B are compared in a 2AFC experiment, we use A$\leftarrowtail$B to denote the percentage of votes for choosing A.

The evaluation result of video stability is shown in Table~\ref{Tab:pixelFDB-video}. First, with either the OFB loss or the P-FDB loss, the model generated more stable videos than without any temporal loss, indicating that these two temporal losses effectively improved video stability. Second, the P-FDB loss won approximately 64.1\%-50.0\%=14.1\% votes against the OFB loss, indicating that the P-FDB loss improved stability more than the OFB loss. Third, there was a non-negligible portion of votes for no temporal loss (21.7\% for the OFB loss and 24.6\% for the P-FDB loss), indicating that in some cases the difference in the stability was not significant.

\begin{table}[t]
    \centering
    \caption{Evaluation results of the SFNs with the P-FDB loss, the OFB loss and without temporal loss (``None'').}
    \label{Tab:pixelFDB}
    \subfigure[Results for video stability.]{
    \label{Tab:pixelFDB-video}
    \begin{tabular}{@{}cccc@{}}
    \hline
                & P-FDB $\leftarrowtail$ OFB & OFB $\leftarrowtail$ None & P-FDB $\leftarrowtail$ None \\ \hline
    Composition    & 70.7\% & 80.0\% & 79.2\%\\
    Feathers       & 62.3\% & 74.3\% & 66.8\%\\
    La Muse        & 57.8\% & 83.5\% & 78.3\%\\
    Starry Night   & 65.8\% & 73.5\% & 77.5\%\\\hline
    Average        & 64.1\% & 78.3\% & 75.4\%\\\hline

    \end{tabular}}

    \subfigure[Results for frame stylization quality.]{
    \label{Tab:pixelFDB-frame}
    \begin{tabular}{@{}cccc@{}}
    \hline
                    & P-FDB $\leftarrowtail$ OFB & OFB $\leftarrowtail$ None & P-FDB $\leftarrowtail$ None\\ \hline
    Composition     & 46.0\% & 40.0\% & 34.2\%\\
    Feathers        & 41.3\% & 39.8\% & 20.3\%\\
    La Muse         & 36.3\% & 55.0\% & 30.3\%\\
    Starry Night    & 30.3\% & 25.0\% & 11.3\%\\\hline
    Average         & 38.6\% & 40.7\% & 24.0\%\\ \hline
    \end{tabular}}
\end{table}

\subsubsection{Frame stylization quality: qualitative results} \label{sec:pixel-quality-qualitative}

We visually inspected the results and found that both the P-FDB loss and the OFB loss showed stylization quality degradation. As shown in Fig.~\ref{fig:sfn_diff_flow_none}, both the P-FDB loss and the OFB loss smoothed out the stylization texture to some extent. This phenomenon is clearly observed in the left example in  Fig.~\ref{fig:sfn_diff_flow_none}.  As indicated by red boxes, with these temporal losses, the sharp stripes presented in the results obtained without any temporal loss were largely suppressed. The stylization quality degradation is not clear in the right example in  Fig.~\ref{fig:sfn_diff_flow_none}.

In principle it is possible that the coefficient for the temporal loss was too large. However, we found that the stylized videos became unstable when we tried to decrease the P-FDB loss coefficient $\lambda_3$.

\subsubsection{Frame stylization quality: quantitative results} \label{sec:pixel-quality-quantitative}

We evaluated the stylization quality of the results obtained with different temporal losses and without any temporal loss by conducting 2AFC experiments as described in~\ref{sec:eval}.
The results are presented in Table~\ref{Tab:pixelFDB-frame}.
Both the P-FDB loss and the OFB loss received significantly fewer votes than no temporal loss, indicating that both temporal losses degraded the stylization quality.
 Compared with no temporal loss, the votes received by the P-FDB loss were fewer than the votes received by the OFB loss, indicating that the P-FDB loss affected the stylization quality more than the OFB loss. A direct comparison between the two temporal losses verified this conclusion. The P-FDB loss received only 38.6\% votes while the OFB loss received 61.4\% votes.

The stylization quality degradation caused by the P-FDB loss may be due to its aggressive requirement to stabilize the video, i.e., making the pixel-to-pixel difference in the stylized video frames similar to that of the original video frames. According to \eqref{eq:temploss} and \eqref{eq:framediff}, when $l=0$,

\begin{equation}
    L_\text{temp} \propto \sum_{t=1}^{T - 1} \norm{(\tilde I_{t+1}-I_{t+1}) - (\tilde I_t-I_t)}_2^2.
\end{equation}

\noindent The minimum is attained if $\tilde I_{t}-I_{t}$ is a constant for all $t$. The stylization image will be close to the original image in this case. Although the style loss counteracts this trend,  the stylization quality may be degraded.

The stylization quality degradation caused by the OFB loss may be due to the complexity of the videos in the DAVIS dataset. According to the official statistics~\cite{Perazzi2016}, 60\% of videos have complicated deformations, 42\% of videos contain fast-moving objects (per-frame object motion is larger than 20 px), and 36\% of videos contain occluded objects and other difficult cases. As mentioned in Section \ref{sec:intro}, these scenarios are challenging for optic flow estimation and will lead to inaccurate optical flow estimation. It is difficult to design a better optic flow estimation algorithm to avoid stylization quality degradation of the OFB loss, because  this task is also quite challenging.

\subsection{Results with the C-FDB loss}\label{sec:combinedFDB}
The previous experiments showed that the P-FDB loss led to style quality degradation. As discussed above, we suspect that the constraint of  enforcing a pixel-level frame difference is too strong. This constraint is relaxed to some extent if we define the FDB loss in the feature space, i.e., $L_\text{ft}$. The training pipeline is illustrated in Fig.~\ref{fig:SFNb}. We note that if $l \ne 0$ in \eqref{eq:temploss} and \eqref{eq:framediff},
\begin{equation}
    L_\text{temp} \propto \sum_{t=1}^{T - 1} \norm{(f_l(\tilde I_{t+1})-f_l(I_{t+1})) - (f_l(\tilde I_t)-f_l(I_t))}_2^2.
\end{equation}
The minimum is attained if $f_l(\tilde I_{t})-f_l(I_{t})$ is a constant for all $t$. However, this does not imply that $\tilde I_{t}-I_{t}$ must be a constant due to the nonlinear transformations (e.g., max pooling) in the neural network from $x$ to $f_l(x)$. Considering that tiny flickering may be neglected by $L_\text{ft}$, we combined $L_\text{ft}$ with $L_\text{pt}$ to form the C-FDB loss $L_\text{ft}$. In this subsection, we present the results of the C-FDB loss.

\subsubsection{Video stability: qualitative results}
\label{sec:comb-stabiltiy-qualitative}
\begin{figure}
    \centering
    \subfigure[Stylized using the ``Composition VII'' image.]{\label{fig:FDB-compare-a}
    \includegraphics[width=1\linewidth]{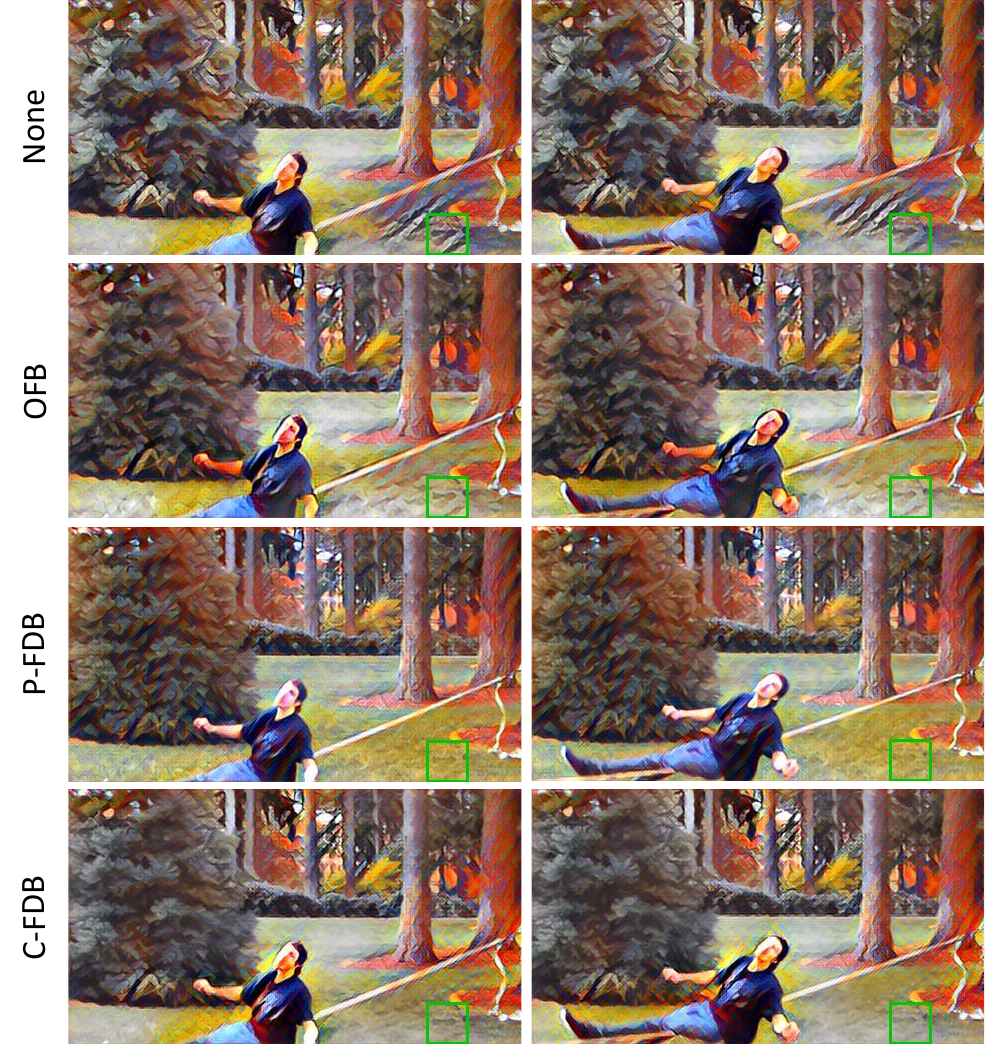}}
    \subfigure[Stylized using the ``Starry Night'' image.]{\label{fig:FDB-compare-b}
    \includegraphics[width=1\linewidth]{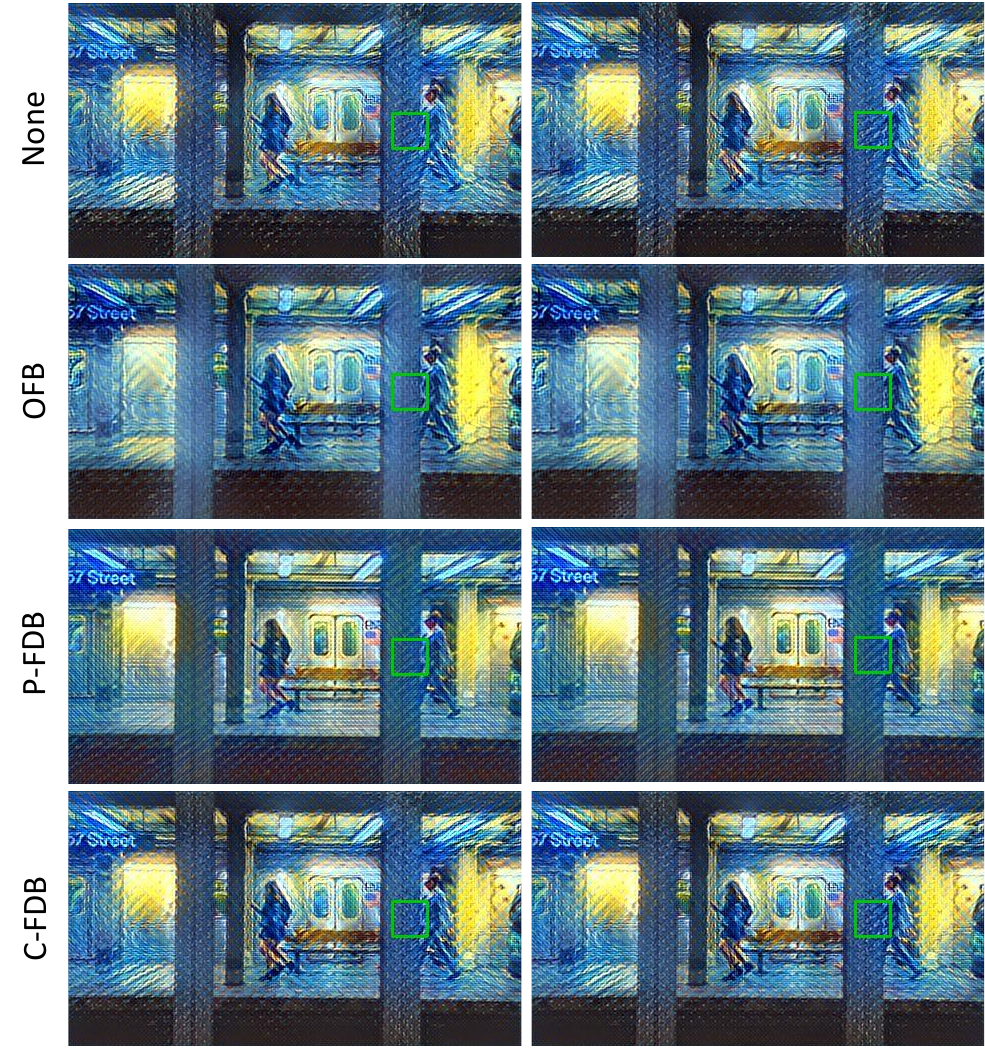}}
    \caption{Stylization results of the SFNs trained without temporal loss, with the OFB loss, the P-FDB loss and the C-FDB loss. See Supplementary Video 5. }
    \label{fig:FDB-compare}
\end{figure}

We compared the results obtained using the SFNs trained with the C-FDB loss, the P-FDB loss, the OFB loss and without any temporal loss.
Several stylized videos are shown in Supplementary Video 5.
Two successive frames stylized by SFNs trained with three temporal losses and without temporal loss are shown in Fig.~\ref{fig:FDB-compare}. Without temporal loss, the two frames displayed large difference as highlighted by green boxes, leading to flickers in the videos. However, the flickers were largely reduced in the results obtained with temporal losses.

\subsubsection{Video stability: quantitative results}
\label{sec:comb-stabiltiy-quantitative}

We conducted 2AFC experiments about video stability with the results listed in Table~\ref{Tab:combinedFDB-video}. First, the C-FDB loss received 72.9\% votes compared with no temporal loss, indicating that the C-FDB loss successfully improved the video stability.
Second, the C-FDB loss received fewer votes than the P-FDB loss, indicating that the stabilization ability of the C-FDB loss was lower than that of the P-FDB loss.
Third, the difference of the votes between the C-FDB loss and the OFB loss was within 2\%, indicating that the two losses had similar capability to stabilize stylized videos.

\begin{table}
    \centering
    \caption{Evaluation results of the SFNs with the C-FDB loss, the P-FDB loss, the OFB loss and without temporal loss (``None'')}
    \label{Tab:combinedFDB}
    \subtable[Results for video stability.]{
    \label{Tab:combinedFDB-video}
    \begin{tabular}{@{}cccc@{}}
    \hline
    Method & C-FDB $\leftarrowtail$ P-FDB & C-FDB $\leftarrowtail$ OFB & C-FDB $\leftarrowtail$ None \\\hline
    Composition  & 37.2\% & 50.5\% & 77.2\% \\
    Feathers     & 48.0\% & 49.7\% & 71.0\% \\
    La Muse      & 39.5\% & 40.5\% & 75.5\% \\
    Starry Night & 34.3\% & 51.5\% & 68.0\% \\\hline
    Average      & 39.8\% & 48.1\% & 72.9\% \\
    \hline
    \end{tabular}}

    \subtable[Results for frame stylization quality.]{
    \label{Tab:combinedFDB-frame}
    \begin{tabular}{@{}cccc@{}}
        \hline
        Method & C-FDB $\leftarrowtail$ P-FDB & C-FDB $\leftarrowtail$ OFB & C-FDB $\leftarrowtail$ None \\ \hline
        Compositions    & 50.2\% & 40.7\% & 50.0\% \\
        Feathers        & 59.8\% & 55.2\% & 36.3\% \\
        La Muse         & 67.5\% & 50.0\% & 47.3\% \\
        Starry Night    & 68.8\% & 57.8\% & 31.0\% \\\hline
        Average         & 61.6\% & 50.9\% & 41.6\% \\
        \hline
    \end{tabular}}
\end{table}

\subsubsection{Frame stylization quality: qualitative results}
\label{sec:comb-quality-qualitative}

 We then compared the stylization quality of the frames obtained with different temporal losses.
 A visual inspection showed that the C-FDB loss and the OFB loss led to richer texture variations in content images than the P-FDB loss.
 In the examples shown in Fig.~\ref{fig:FDB-compare}, the results obtained with the C-FDB loss and the OFB loss had similar stylization quality to the result obtained without temporal loss but displayed clearly higher quality than the result obtained with the P-FDB loss. In fact, in Fig.~\ref{fig:FDB-compare-a}, the faces in the 1st, 2nd and 4th rows have richer colors than the faces in the 3rd row. In  Fig.~\ref{fig:FDB-compare-b}, the bodies of the female pedestrian in the 2nd and the 4th rows have more similar colors  than the 3rd row to the bodies in the 1st row.


\subsubsection{Frame stylization quality: quantitative results}
\label{sec:comb-quality-quantitative}

 Quantitative evaluations were performed in the same manner as that used to compare video stability (see Section \ref{sec:comb-stabiltiy-quantitative}). The C-FDB loss was compared with the P-FDB loss, the OFB loss, and no temporal loss.
The results are shown in Table~\ref{Tab:combinedFDB-frame}. Compared with no temporal loss, the C-FDB loss received 41.6\% votes. Compared with the P-FDB loss, the C-FDB loss received 61.6\% votes. Compared with the OFB loss, the C-FDB loss received 50.9\% votes. The results indicated that the stylization quality of the C-FDB loss matched that of the OFB loss but was higher than that of the P-FDB loss.

\subsection{Results of the RNN}\label{sec:exp-RNN}

We trained two RNNs with the C-FDB loss and the OFB loss, respectively. As a baseline model, the SFN trained without any temporal loss was also included in comparison. Comparison with this baseline model is meaningful because this model was used as the initial model for training the RNNs (see Section~\ref{subsec:implementation}). We performed 2AFC experiments with the same setting as described in Section \ref{sec:eval}.

\subsubsection{Qualitative results}

  Several example videos stylized by the RNNs are shown in Supplementary Video 6. Fig.~\ref{fig:rnn_flow_comb_none} shows two successive frames from different videos. For video stability, as expected, the two frames obtained by the SFN trained without temporal loss were inconsistent as highlighted by blue boxes. In Fig.~\ref{fig:rnn_flow_comb_none_feathers}, the shape and the number of stripes are clearly different in the two frames. In Fig.~\ref{fig:rnn_flow_comb_none_composition}, the black-white texture is sharper in the second frame. The two frames output by the RNN trained with the OFB loss or the C-FDB loss were more consistent. In fact, the videos produced by the two RNNs were more stable (see Supplementary Video 6).
In terms of stylization quality, the results of the two RNNs were hardly distinguishable.

\begin{figure}
\centering
    \subfigure[Stylized using the ``Feathers'' image]{\label{fig:rnn_flow_comb_none_feathers}
        \includegraphics[width=1\linewidth]{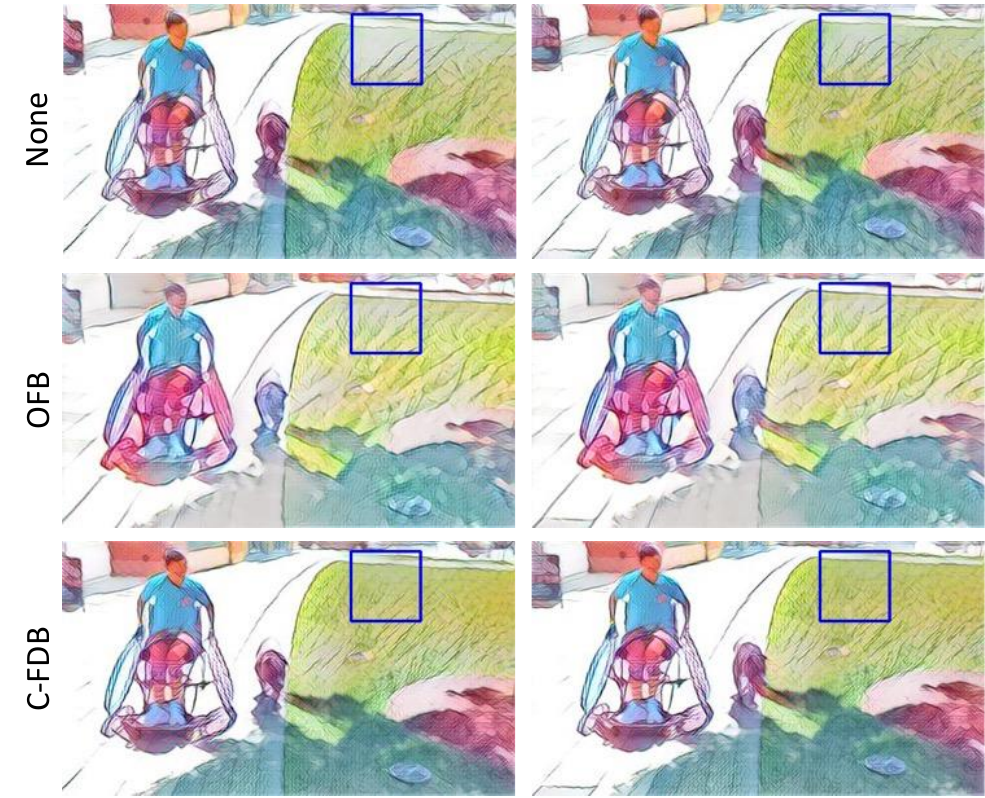}
    }
    \subfigure[Stylized using the ``Composition VII'' image]{\label{fig:rnn_flow_comb_none_composition}
        \includegraphics[width=1\linewidth]{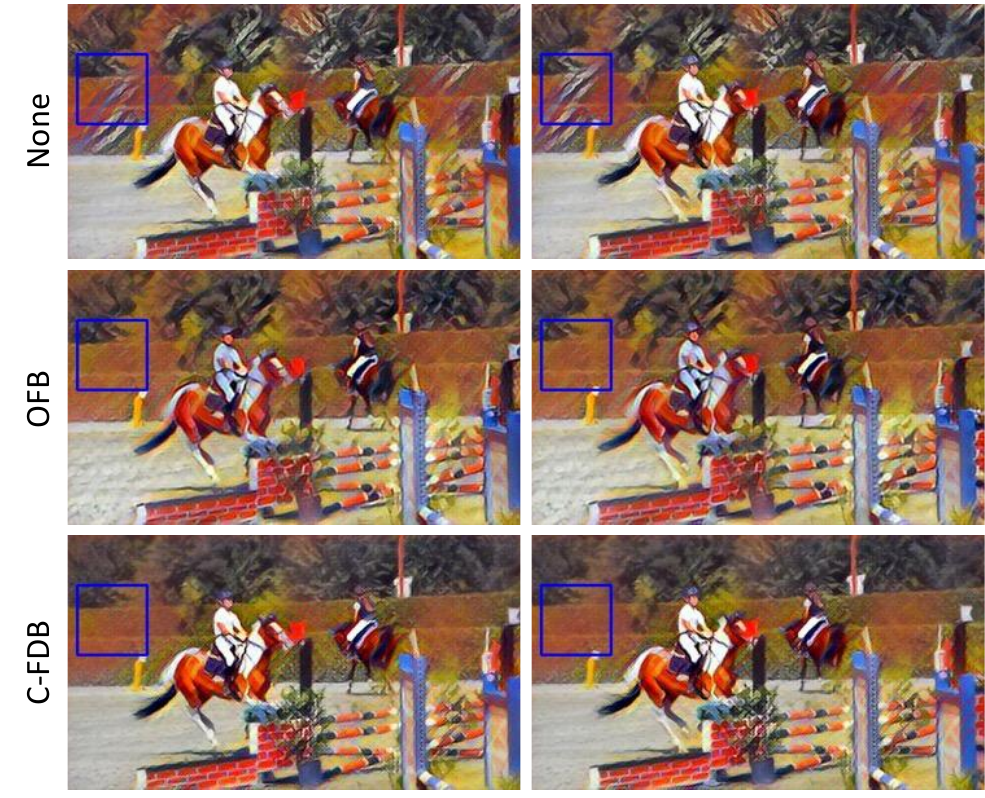}
    }
    \caption{Stylization results of the SFN trained without any temporal loss (top), and the results of the RNNs trained with the OFB loss (middle) and the C-FDB loss (bottom). See Supplementary Video 6. }
    \label{fig:rnn_flow_comb_none}
\end{figure}

\subsubsection{Quantitative results}

\begin{table}
    \caption{Evaluation results of the RNNs trained with the C-FDB loss, the OFB loss and the results of the SFN trained without any temporal loss (``None''). }
    \label{Tab:RNN}
    \centering
    \subtable[Results for video stability.]{
        \label{Tab:RNN-video}
        \begin{tabular}{@{}ccc@{}}
        \hline
        Method & C-FDB $\leftarrowtail$ OFB & C-FDB $\leftarrowtail$ None  \\\hline
        Composition   & 74.7\% & 70.0\% \\
        Feathers      & 59.3\% & 67.3\% \\
        La Muse       & 49.3\% & 75.5\% \\
        Starry Night  & 58.3\% & 71.0\% \\\hline
        Average       & 60.4\% & 72.2\% \\\hline
        \end{tabular}
    }
    \subtable[Results for frame stylization quality.]{
        \label{Tab:RNN-frame}
        \begin{tabular}{@{}ccc@{}}
            \hline
            Method & C-FDB $\leftarrowtail$ OFB & C-FDB $\leftarrowtail$ None \\\hline
            Composition   & 71.2\% & 48.2\% \\
            Feathers      & 59.8\% & 36.8\% \\
            La Muse       & 53.5\% & 51.8\% \\
            Starry Night  & 57.8\% & 36.3\% \\\hline
            Average       & 60.6\% & 43.2\% \\\hline
        \end{tabular}
    }
\end{table}

Tables~\ref{Tab:RNN-video} and ~\ref{Tab:RNN-frame} show the 2AFC results for video stability and frame stylization quality, respectively. For stability, the C-FDB loss received 72.2\% votes compared with no temporal loss and 60.4\% votes compared with the OFB loss. For stylization quality, the C-FDB received 43.2\% votes compared to the method with no temporal loss and 60.6\% votes compared to the method with the OFB loss. These results suggested that the C-FDB loss was a better choice than the OFB loss for training the RNN.

Compared with the SFN trained without any temporal loss, the RNN trained with the C-FDB loss (Table~\ref{Tab:RNN}) and the SFN trained with the C-FDB loss (Table~\ref{Tab:combinedFDB}) performed equally well in terms of either video stability or frame stylization quality. However, the results of the comparison between the C-FDB loss and the OFB loss showed that the C-FDB loss was better on the RNN than on the SFN. In fact, on the RNN, the C-FDB received 60.4\% votes for video stability  compared with the OFB loss (Table~\ref{Tab:RNN-video}), but this number was 48.1\% on the SFN (Table~\ref{Tab:combinedFDB-video}). On the RNN, the C-FDB received 60.6\% votes for frame stylization quality when compared with the OFB loss (Table~\ref{Tab:RNN-frame}), but this value was 50.9\% on the SFN (Table~\ref{Tab:combinedFDB-frame}).

\subsection{Summary of the experimental results}

We summarize the experimental results presented in this section:

\begin{itemize}
 \item On the SFN, the P-FDB loss produced better stability results than the OFB loss, but both losses suffered from stylization quality degradation when compared with no temporal loss.
 \item On the SFN, the C-FDB loss was better than the P-FDB loss in terms of stylization quality.
 \item On the SFN, the C-FDB loss was comparable to the OFB loss in terms of stability and stylization quality.
 \item On the RNN, the C-FDB loss was better than the OFB loss in terms of stability and stylization quality.
\end{itemize}
We note that both the SFN and the RNN were significantly faster in training with the C-FDB loss than with the OFB loss, because the C-FDB loss did not require the time-consuming estimation of optical flows for videos.

\section{Concluding Remarks}\label{sec:conclusions}

The cutting-edge video stylization models usually adopt OFB methods to stabilize frames. However, the calculation of optical flow is expensive and challenging in complex scenarios. Since stability is encoded in the frame difference in the original video, we propose to stabilize the frames by requiring the output frame difference to be close to the input frame difference. The FDB loss is formulated and can be calculated either in the pixel space, or in the feature space of certain deep neural networks, or in both spaces. This technique is simple yet effective. We compared the performances of the proposed loss and the popular OFB temporal loss on two typical video stylization networks. Extensive human behavior experiments showed that the proposed loss was comparable to the OFB loss on one network and better than the OFB loss on the other network in terms of stability and stylization quality of the videos.

The proposed FDB loss can be applied to most video stylization models that use the OFB loss. In addition, it may have applications beyond video stylization. Indeed, in any task that requires temporal coherence between the video frames while the original stable video is known, e.g., color grading \cite{bonneel2013example} and tonal stabilization \cite{farbman2011tonal}, the proposed frame difference-based method or its extension may be a good substitute for the optic flow-based methods. Using the OFB loss means that users must understand and correctly set up the optic flow estimation pipeline, requiring an intense  effort in practice. By contrast, the C-FDB loss can be applied to new models similar to any other loss functions used to train neural networks. It does not require time-consuming preprocessing. We believe that the proposed method will be widely applied in video processing.

\section*{Acknowledgments}
The authors would like to thank Yueqiao Li and Jian Wu for helping to set up the experiments.
This work was supported in part by the National Natural Science
Foundation of China under Grant 62061136001, Grant
61836014, Grant U19B2034 and Grant 61620106010.

\bibliographystyle{ieeetran}
\bibliography{references}

\end{document}